\documentclass[conference]{IEEEtran}
\IEEEoverridecommandlockouts

\usepackage{cite}
\usepackage{amsmath,amssymb,amsfonts}
\usepackage{algorithmic}
\usepackage{graphicx}
\usepackage{textcomp}
\usepackage{url}
\usepackage{xcolor}
\usepackage{booktabs}
\def\BibTeX{{\rm B\kern-.05em{\sc i\kern-.025em b}\kern-.08em
    T\kern-.1667em\lower.7ex\hbox{E}\kern-.125emX}}
\begin{document}

\title{Revisiting Gradient Staleness: Evaluating Distance
Metrics for Asynchronous Federated Learning
Aggregation\\

}

\author{\IEEEauthorblockN{Patrick Wilhelm}
\IEEEauthorblockA{\textit{BIFOLD} \\
\textit{Technische Universität Berlin}\\
Berlin, Germany \\
patrick.wilhelm@tu-berlin.de}
\and
\IEEEauthorblockN{Odej Kao}
\IEEEauthorblockA{\textit{Technische Universität Berlin} \\
Berlin, Germany \\
}
\and

}

\maketitle

\begin{abstract}
In asynchronous federated learning (FL), client devices send updates to a central server at varying times based on their computational speed, often using stale versions of the global model. This staleness can degrade the convergence and accuracy of the global model. Previous work, such as AsyncFedED, proposed an adaptive aggregation method using Euclidean distance to measure staleness. In this paper, we extend this approach by exploring alternative distance metrics to more
accurately capture the effect of gradient staleness. We integrate these metrics into the aggregation process and evaluate their impact on convergence speed, model performance, and training stability under heterogeneous clients and non-IID data settings. Our results demonstrate that certain metrics lead to more robust and efficient asynchronous FL training, offering a stronger foundation for practical deployment.
\end{abstract}


\section{Introduction}

Federated Learning (FL) has emerged as a promising paradigm for privacy-preserving, decentralized model training across distributed client devices. By enabling local model updates without sharing raw data, FL addresses key privacy and communication challenges, especially in edge computing and mobile scenarios. However, traditional synchronous FL approaches require coordinated participation from multiple clients per training round. This makes them vulnerable to system heterogeneity, client stragglers, and network latency, limiting their scalability and practicality in real-world deployments. \cite{b8}

To overcome these limitations, asynchronous Federated Learning (AFL) has gained increasing attention. In AFL, the server updates the global model whenever it receives a client update, allowing clients to operate independently based on their availability. While this improves efficiency and reduces idle time, it introduces a critical challenge: gradient or model staleness. Client updates are often computed on outdated versions of the global model, leading to reduced convergence speed, degraded accuracy, and training instability—especially in non-IID settings. \cite{b23, b20}

Several works have addressed staleness through adaptive aggregation strategies, most notably AsyncFedED \cite{b11}, which uses Euclidean distance between local and global models to weight client updates. While effective, this approach assumes that a single geometric distance metric is sufficient to quantify staleness. However, model divergence is multi-faceted: updates may differ in direction (e.g., angular similarity), in statistical properties (e.g., covariance structure), or in distributional characteristics (e.g., under non-IID conditions). A scalar distance metric may not capture these nuances. \cite{b21, b22}

In this work, we extend the current understanding of staleness in AFL by evaluating a broader class of distance metrics as tools to measure and respond to gradient staleness. We integrate these metrics into the aggregation process and assess their impact on model convergence, training stability, and performance in diverse asynchronous FL scenarios.

Our contributions are as follows:

\begin{itemize}
\item We systematically analyze multiple distance metrics for quantifying gradient staleness in asynchronous FL
\item We perform extensive experiments under varying degrees of system heterogeneity, demonstrating that certain distance metrics yield more robust and efficient training than traditional approaches.
\end{itemize}

By offering a principled and flexible way to quantify and mitigate staleness, this work moves AFL closer to practical deployment in heterogeneous, real-world environments.

\section{Related Work}

\begin{table*}[h!]
\caption{Final Test Accuracy (Mean ± Std) for Each Staleness Metric Across Availability Scenarios [\%]}
\begin{center}
\begin{tabular}{|l|c|c|c|}
\hline
\textbf{Staleness Metric} & \textbf{Low} & \textbf{Medium} & \textbf{High} \\
\hline
Bregman        & 0.40 ± 0.02 & 0.40 ± 0.02 & 0.39 ± 0.01 \\
\hline
Euclidean      & 0.41 ± 0.02 & 0.43 ± 0.01 & 0.44 ± 0.01 \\
\hline
Fisher         & 0.34 ± 0.02 & 0.34 ± 0.01 & 0.35 ± 0.01 \\
\hline
Manhattan      & 0.24 ± 0.00 & 0.24 ± 0.00 & 0.25 ± 0.02 \\
\hline
Cosine         & 0.20 ± 0.02 & 0.27 ± 0.02 & 0.21 ± 0.03 \\
\hline
KL-divergence  & 0.28 ± 0.03 & 0.17 ± 0.09 & 0.25 ± 0.05 \\
\hline
Hellinger      & 0.24 ± 0.07 & 0.25 ± 0.07 & 0.21 ± 0.06 \\
\hline
\end{tabular}
\label{tab:final_accuracy}
\end{center}
\end{table*}

Distributed Machine Learning (DML) encompasses a set of techniques designed to enable scalable and efficient training of machine learning models across multiple computational units \cite{b2}. Early approaches, such as \textit{"MapReduce for Machine Learning on Multicore"} \cite{b3}, demonstrated how computation could be parallelized across CPU cores to accelerate the training of classical algorithms like Support Vector Machines (SVMs) and neural networks.

A cornerstone of distributed training is the use of Stochastic Gradient Descent (SGD) for optimizing model parameters. The basic SGD update rule is given by:

\begin{equation}
x_{t+1} = x_t - \eta_t g_t
\end{equation}

where \(x_t\) denotes the model parameters at iteration \(t\), 
\(\eta\) is the learning rate, and \(g_t\) is the stochastic gradient computed from a data mini-batch.

A natural strategy to parallelize SGD is to compute gradients independently across multiple workers and replace  \(g_t\) in Equation (1) with the average of these gradients \cite{b7}. This technique, often coupled with mini-batching, forms the foundation of data-parallel training.

In shared-memory environments, asynchronous variants such as \textit{Hogwild!} \cite{b5} and \textit{Buckwild!} \cite{b6} have shown that lock-free updates can be effective, despite the presence of gradient staleness. These approaches leverage the sparsity of updates to mitigate interference and achieve high-throughput training.

\textbf{From Data Parallelism to Federated Learning}\\
A standard approach to distributed training is data parallelism, where the training dataset is partitioned among multiple nodes. Each node maintains a full replica of the model and performs local updates using its data shard. This strategy is effective in both tightly coupled environments (e.g., GPU clusters) and loosely coupled systems spanning multiple machines. Data parallelism addresses scalability bottlenecks in single-node training and is foundational to both distributed deep learning and federated learning frameworks. \cite{b4}

Federated Learning (FL) extends distributed training to settings where data is inherently decentralized, such as on mobile devices or edge sensors. FL allows clients to collaboratively train a global model without sharing raw data, thereby preserving privacy and reducing communication overhead. Synchronous protocols like FedAvg \cite{b8} aggregate client updates periodically. However, their reliance on full client participation and synchronization can lead to straggler issues and inefficiencies in real-world deployments.

To overcome these limitations, Asynchronous Federated Learning (AsyncFL) has emerged as a promising paradigm \cite{b9, b10}. In AsyncFL, the server updates the global model upon receiving individual client updates, without waiting for all participants. While this improves responsiveness and utilization, it introduces the problem of gradient staleness, where updates are computed based on outdated global models. This staleness can lead to slower convergence and degraded model quality.

To address this, several adaptive aggregation strategies have been proposed. Notably, AsyncFedED \cite{b11} proposed a Euclidean distance-based adaptive weighting mechanism that downweights stale updates based on the distance between the client's local model and the current global model, as well as the number of local training epochs. This approach demonstrated improved convergence in asynchronous FL setups and provided a practical method for quantifying staleness.

Other related methods include temporal discounting \cite{b27}, staleness-aware gradient clipping \cite{b12}, and importance-aware update scheduling \cite{b13}. However, most existing work either assumes a simple scalar staleness model (e.g., timestamp lag) or relies on a fixed distance metric like Euclidean distance, which may not capture the semantic or statistical divergence between models, especially under non-IID data and heterogenous environments.

\textbf{Mathematical Distance Metrics}\\
Mathematical distance metrics are broadly categorized by their underlying geometric and statistical foundations, each offering different capabilities for capturing divergence in high-dimensional optimization problems. Euclidean geometry provides the most direct measures, such as the L2 (Euclidean) and L1 (Manhattan) distances, which treat model parameters as flat vectors in metric space. While computationally efficient, such metrics may oversimplify the underlying model space, especially in settings involving statistical structure or high heterogeneity.

Riemannian geometry generalizes Euclidean space by introducing curvature through metrics like the Fisher-Rao distance, which considers the statistical manifold induced by probability distributions \cite{b17}. This is relevant when gradient updates are viewed as samples from evolving distributions, where curvature-aware measures can better represent their divergence.

In information geometry, divergences such as the Kullback-Leibler (KL) divergence \cite{b14}, Jensen-Shannon divergence \cite{b15}, and Bregman divergences \cite{b16} quantify informational differences between probabilistic representations. These metrics are particularly suited for measuring how "outdated" an update is, in terms of the entropy or information content it carries compared to the current model.

Optimal transport geometry, including the Wasserstein distance and Earth Mover’s Distance (EMD), offers another perspective by treating updates as distributions and computing the minimal cost to morph one into another \cite{b19}. These measures naturally account for structural shifts and mass displacement, which can occur in asynchronous FL when updates are delayed and clients train on non-IID data.

Finally, statistical distances like the Mahalanobis distance \cite{b24}, cosine similarity, or Hellinger distance incorporate variance \cite{b18}, directional alignment, or probabilistic overlap into the distance computation. These measures can provide more nuanced assessments of how a delayed gradient update differs from the current optimization trajectory, beyond simple magnitude.

In the context of asynchronous optimization, the choice of distance metric directly affects the ability to quantify and correct for staleness. Different metrics offer trade-offs between computational complexity, geometric fidelity, and statistical interpretability. As such, a principled understanding of the taxonomy of distance measures is essential for designing robust and adaptive aggregation strategies in asynchronous federated learning systems.
By integrating selected metrics from this broader mathematical landscape into asynchronous aggregation schemes, we aim to enable finer-grained, geometry-aware staleness modeling. Our experiments demonstrate that these alternative metrics can significantly enhance convergence stability and robustness in heterogeneous settings.

\section{Methodology}
\label{sec:method}

We investigate the impact of various mathematical distance metrics on gradient staleness modeling in asynchronous Federated Learning (FL), building upon the AsyncFedED \cite{b11} framework. While AsyncFedED uses Euclidean distance to estimate the staleness of client updates, our work systematically compares this with alternative metrics to explore their effect on model convergence, robustness, and performance.

We modify the original AsyncFedED staleness estimator to support multiple distance metrics. Specifically, we generalize the numerator of the staleness function:

\begin{equation}
\gamma(i, \tau) = \frac{D(x_t, x_{t-\tau})}{\left\| \Delta_i(x_{t-\tau}, K) \right\|_2}
\end{equation}

Where \(D\) is a chosen distance function. The denominator remains the L2 -norm of the client's update, preserving the original intuition that larger updates indicate lower staleness.
Here, \(x_t\) denotes the current global model parameters at the time the server receives the client update, while \(x_{t-\tau}\) represents the global model parameters at the time the client began its local training, \(\tau\) steps earlier. Thus, the numerator \(D(x_t, x_{t - \tau})\) captures how much the global model has changed during the client's training period, effectively measuring the "staleness" of the update due to asynchrony.

Each metric is integrated into the adaptive global learning rate calculation:

\begin{equation}
\eta_{g,i} = \frac{\lambda}{\gamma(i, \tau) + \epsilon}
\end{equation}

where:
\begin{itemize}
    \item $\eta_{g,i}$ is the global learning rate applied to client $i$'s update,
    \item $\lambda$ is a tunable scaling factor,
    \item $\gamma(i, \tau)$ is the staleness of the update,
    \item $\epsilon$ is a small constant to avoid division by zero.
\end{itemize}

and is also used in guiding the adjustment of local training epochs per client to stabilize staleness across the network.

To evaluate the impact of distance and divergence measures in asynchronous federated learning (AFL), we design a controlled experimental setup where stale client updates are aggregated using different distance-aware strategies. Our goal is to understand how various metrics influence convergence, stability, and robustness under non-IID and asynchronous conditions.

\textbf{Metric Solution:}
We select six representative metrics spanning different geometric and functional categories. This selection balances computational feasibility with theoretical diversity, see Table \ref{tab:distance-metrics}.

\begin{table*}[t]
\centering
\caption{Selected distance/divergence metrics evaluated in our study}
\label{tab:distance-metrics}
\begin{tabular}{|l|l|l|}
\hline
\textbf{Metric} & \textbf{Type} & \textbf{Rationale} \\

\hline
L2 Distance & Euclidean norm &  Measures magnitude of update deviation. \\
\hline
L1 Manhattan Distance & Euclidean norm (L1) & Measures total coordinate-wise deviation. \\
\hline
Cosine Distance & Directional similarity & Captures alignment/conflict in gradient directions. \\
\hline
Bregman Divergence & Information-theoretic (asymmetric) & Measures information loss using convex functions. \\
\hline
Hellinger Distance & Probabilistic (symmetric) & Useful for comparing probability distributions. \\
\hline
KL-Divergence & Information-theoretic (asymmetric) & Measures relative entropy \\
\hline
Fisher Information Distance & Riemannian geometry & Captures curvature of the loss surface differences. \\
\hline
\end{tabular}
\end{table*}

\textbf{Simulation Framework:}
We simulate asynchronous Federated Learning (FL) using a customized setup based on the Flower framework \cite{b25}. The server asynchronously receives model updates from clients and performs global aggregation without waiting for all clients to finish. Each client performs local training and then waits for a randomly sampled delay before sending its model update to the server, simulating real-world heterogeneity in compute and network conditions. Data heterogeneity is modeled using the Diriclet distribution with an alpha of 0.5 \cite{b26}.

\textbf{Modeling Asynchrony:}
To emulate varying degrees of client-side latency and participation gaps, we introduce random wait functions drawn from a clipped normal distribution:

\[
\text{delay} \sim \text{clip}(\mathcal{N}(\mu, \sigma^2), 0, \text{max})
\]
\\
 
\begin{table*}[t]
\centering
\caption{Asynchrony Scenarios used in Experiments}
\label{tab:asynchrony_scenarios}
\begin{tabular}{|l|c|c|c| p{7.0cm}|}
\hline
\textbf{Scenario} & \textbf{Mean ($\mu$)} & \textbf{Std. Dev. ($\sigma$)} & \textbf{Max Delay (s)} & \textbf{Description} \\
\hline
Low Asynchrony & 1.0 & 0.5 & 3 & Emulates environments with uniform devices and reliable network. \\
\hline
Moderate Asynchrony & 3.0 & 1.0 & 6 & Represents typical real-world heterogeneity in mobile and edge devices. \\
\hline
High Asynchrony & 5.0 & 2.5 & 10 & Simulates scenarios with frequent stragglers and inconsistent participation. \\
\hline
\end{tabular}
\end{table*}

This delay affects the staleness of each client’s update at the time of server aggregation. Table \ref{tab:asynchrony_scenarios} shows the different scanrios we are evaluating.

\section{Evaluation}
\label{sec:experiments}
\subsection{Experimental Setup}

To ensure transparency and reproducibility, we detail the dataset, simulation environment, and evaluation protocol used in our experiments.
\textbf{Computer Vision:}\\
We use the Fashion-MNIST dataset, partitioned in a non-IID manner using a Dirichlet distribution with concentration parameter \(\alpha\) = 0.5. This setup introduces statistical heterogeneity among clients, reflecting realistic federated learning conditions.

The global model is a lightweight convolutional neural network (CNN), adapted from the PyTorch "60-Minute Blitz" tutorial. It consists of two convolutional layers followed by max-pooling, and three fully connected layers.

\textbf{Text Prediction:}

For next character prediction we used the Shakespeare dataset and the LSTM Network introduced in 
\cite{b26}.

In both experiments we simulate 20 clients with asynchronous availability profiles categorized into three types: \textit{low}, \textit{medium} and \textit{high} availability. These profiles reflect real-world scenarios where client participation may be irregular due to energy or network constraints.

We measure Top-1 accuracy over a fixed wall-clock time of 300 seconds (5 minutes), indirectly bounding the computational cost of each metric. This evaluation framework captures the trade-off between model performance and time-constrained execution.
To ensure statistical robustness, each experimental configuration is repeated \textit{N = 10} times, and results are averaged.

\subsection{Learning Curves over Wall-Clock Time}
\subsubsection{Computer Vision}

\begin{figure}[h]
\centering
\includegraphics[width=0.5\textwidth]{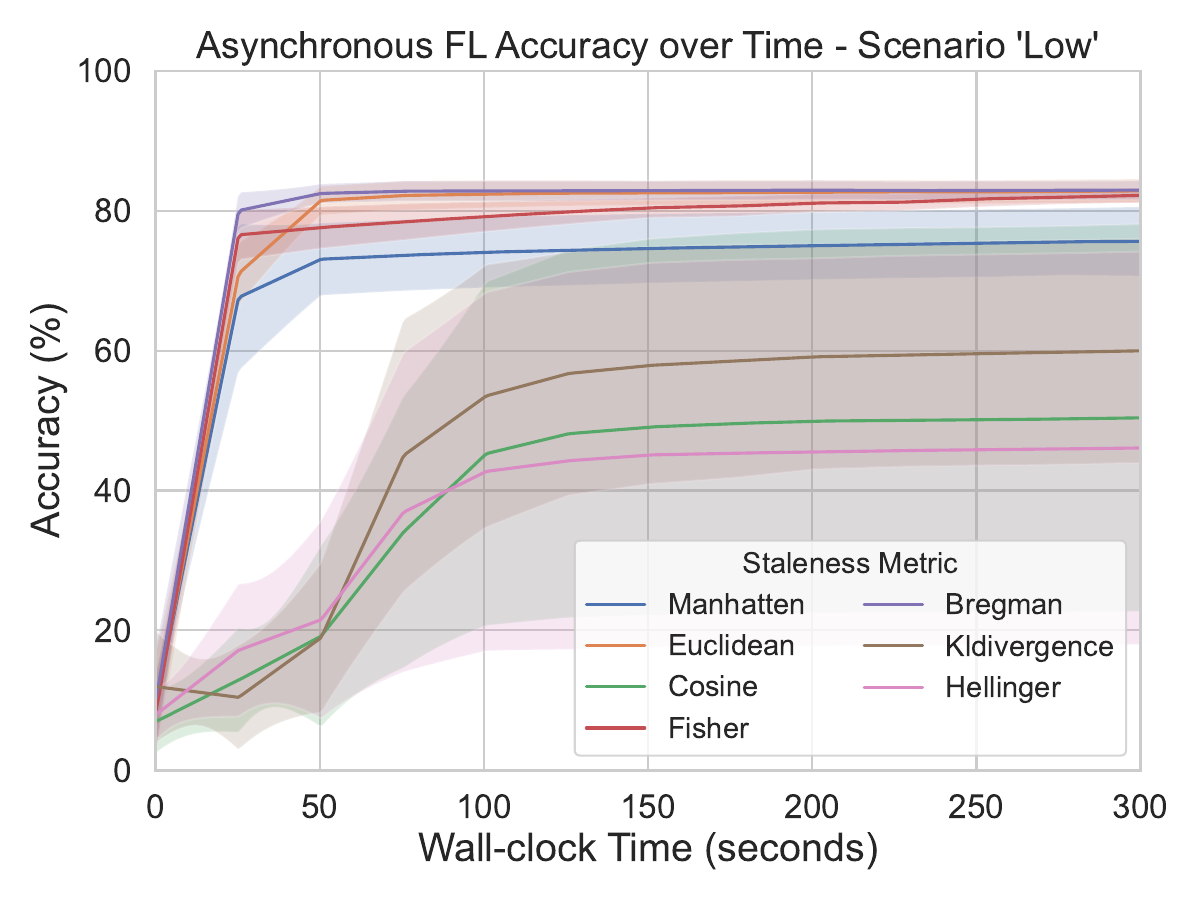}
\caption{Image classification using an CNN Model. Test accuracy over wall-clock time for the \textit{Low} client heterogeneity scenario.}
\label{fig:accuracy_low}
\end{figure}

\begin{figure}[h]
\centering
\includegraphics[width=0.5\textwidth]{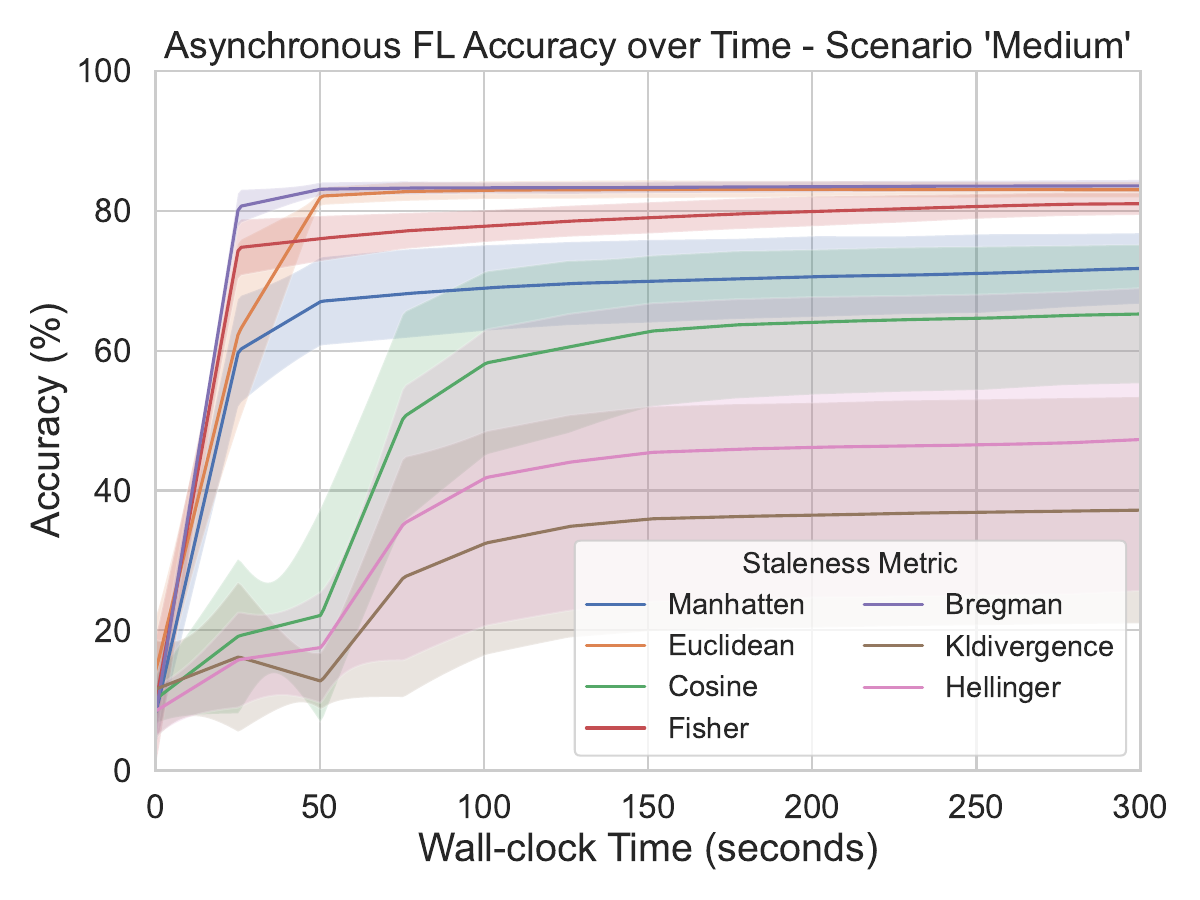}
\caption{Image classification using an CNN Model. Test accuracy over wall-clock time for the \textit{Medium} client heterogeneity scenario.}
\label{fig:accuracy_medium}
\end{figure}

\begin{figure}[h]
\centering
\includegraphics[width=0.5\textwidth]{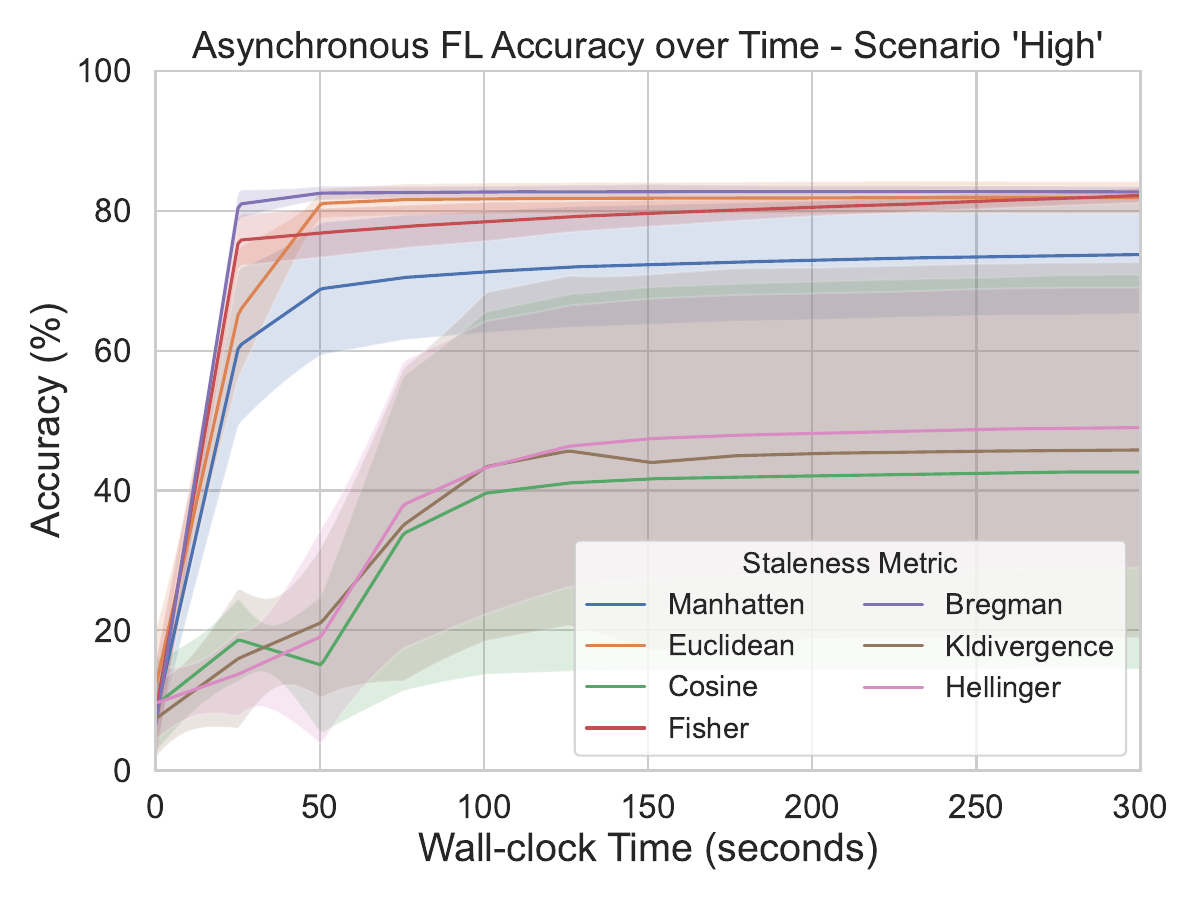}
\caption{Image classification using an CNN Model. Test accuracy over wall-clock time for the \textit{High} client heterogeneity scenario.}
\label{fig:accuracy_high}
\end{figure}

Figures \ref{fig:accuracy_low}, \ref{fig:accuracy_medium} and \ref{fig:accuracy_high} illustrate the progression of test accuracy over wall-clock time across three levels of client availability (Low, Medium, and High). These results are supported by Table \ref{tab:final_accuracy}, which summarizes the final accuracy (mean ± standard deviation) achieved under each scenario for the different staleness metrics.

Across all settings, Bregman divergence consistently yields the highest final test accuracy, demonstrating both early convergence and stable performance, regardless of the level of client asynchrony. Its robustness is particularly evident under high staleness, where other metrics degrade more severely.

In the Low availability scenario (Figure \ref{fig:accuracy_low}), the final accuracies reinforce the visual trends: Euclidean (82.86 ±1.68) and Fisher (82.22 ±1.05) closely follow Bregman, while Manhattan lags slightly behind at 75.63 (±4.93). Information-theoretic metrics such as KL-divergence (59.98 ±16.01), Cosine (50.40 ±27.69), and Hellinger (46.09 ±28.08) show significantly lower performance and higher variance, indicating sensitivity to high staleness and unstable gradient updates.

In the Medium scenario (Figure \ref{fig:accuracy_medium}), performance trends are largely consistent, with Bregman maintaining its lead, and divergence-based methods continuing to underperform, especially in terms of stability.

In the High availability scenario (Figure \ref{fig:accuracy_high}), the gap between the top-performing metrics narrows. Here, Fisher slightly outperforms Euclidean, corroborating the hypothesis that richer curvature-based distances become more beneficial when staleness effects are increased. Nevertheless, Bregman still slightly outperforms both with 82.70 (±0.78), affirming its versatility across all operating regimes.

Interestingly, Euclidean and Bregman exhibit similar convergence profiles across all scenarios, both in the learning curves and final accuracy. This can be attributed to their shared foundation in convex optimization. Many Bregman divergences, such as the squared Euclidean distance, arise from strictly convex functions and measure similar geometric properties of the gradient space. However, Bregman’s additional flexibility, such as its asymmetric structure and curvature sensitivity, may allow it to adapt better in highly asynchronous environments, accounting for its consistent edge.

On the other hand, KL-divergence and Hellinger distances perform poorly, especially under low availability, where their final accuracies fall well below 50 and show large standard deviations. These metrics’ sensitivity to small distributional changes and asymmetry likely make them less robust in the presence of stale or noisy updates.

\begin{table*}[h!]
\caption{Final Test Accuracy (Mean ± Std) for Each Staleness Metric Across Availability Scenarios [\%]}
\begin{center}
\begin{tabular}{|l|c|c|c|}
\hline
\textbf{Staleness Metric} & \textbf{Low} & \textbf{Medium} & \textbf{High} \\
\hline
Bregman        & 82.96 ± 1.35 & 83.57 ± 0.85 & 82.70 ± 0.78 \\
\hline
Euclidean      & 82.86 ± 1.68 & 83.01 ± 1.10 & 81.90 ± 2.29 \\
\hline
Fisher         & 82.22 ± 1.05 & 81.02 ± 1.61 & 82.21 ± 1.04 \\
\hline
Manhattan      & 75.63 ± 4.93 & 71.77 ± 5.04 & 73.75 ± 8.41 \\
\hline
Cosine         & 50.40 ± 27.69 & 65.25 ± 9.91 & 42.65 ± 28.15 \\
\hline
KL-divergence  & 59.98 ± 16.01 & 37.19 ± 16.19 & 45.80 ± 26.85 \\
\hline
Hellinger      & 46.09 ± 28.08 & 47.30 ± 21.65 & 49.00 ± 20.01 \\
\hline
\end{tabular}
\label{tab:final_accuracy}
\end{center}
\end{table*}

In conclusion, the evaluation clearly shows that Bregman divergence offers a compelling balance of stability, convergence speed, and final accuracy, making it a strong candidate for staleness-aware optimization in asynchronous federated learning. Its consistent performance across client availability regimes underscores its robustness and general applicability.

\subsubsection{Text Prediction}

\begin{figure}[h]
\centering
\includegraphics[width=0.5\textwidth]{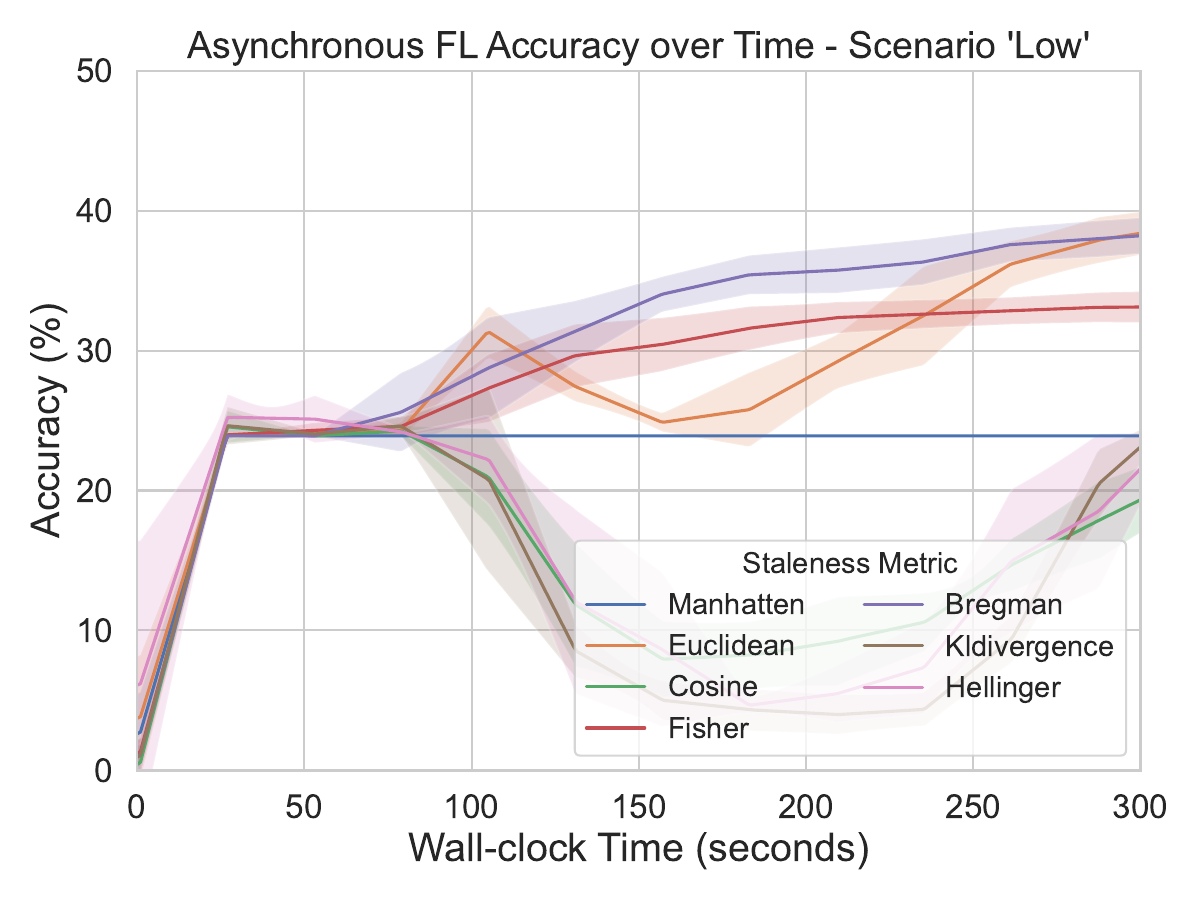}
\caption{Next Character Prediction using an LSTM Model. Test accuracy over wall-clock time for the \textit{Low} client heterogeneity scenario.}
\label{fig:lstm_accuracy_low}
\end{figure}

\begin{figure}[h]
\centering
\includegraphics[width=0.5\textwidth]{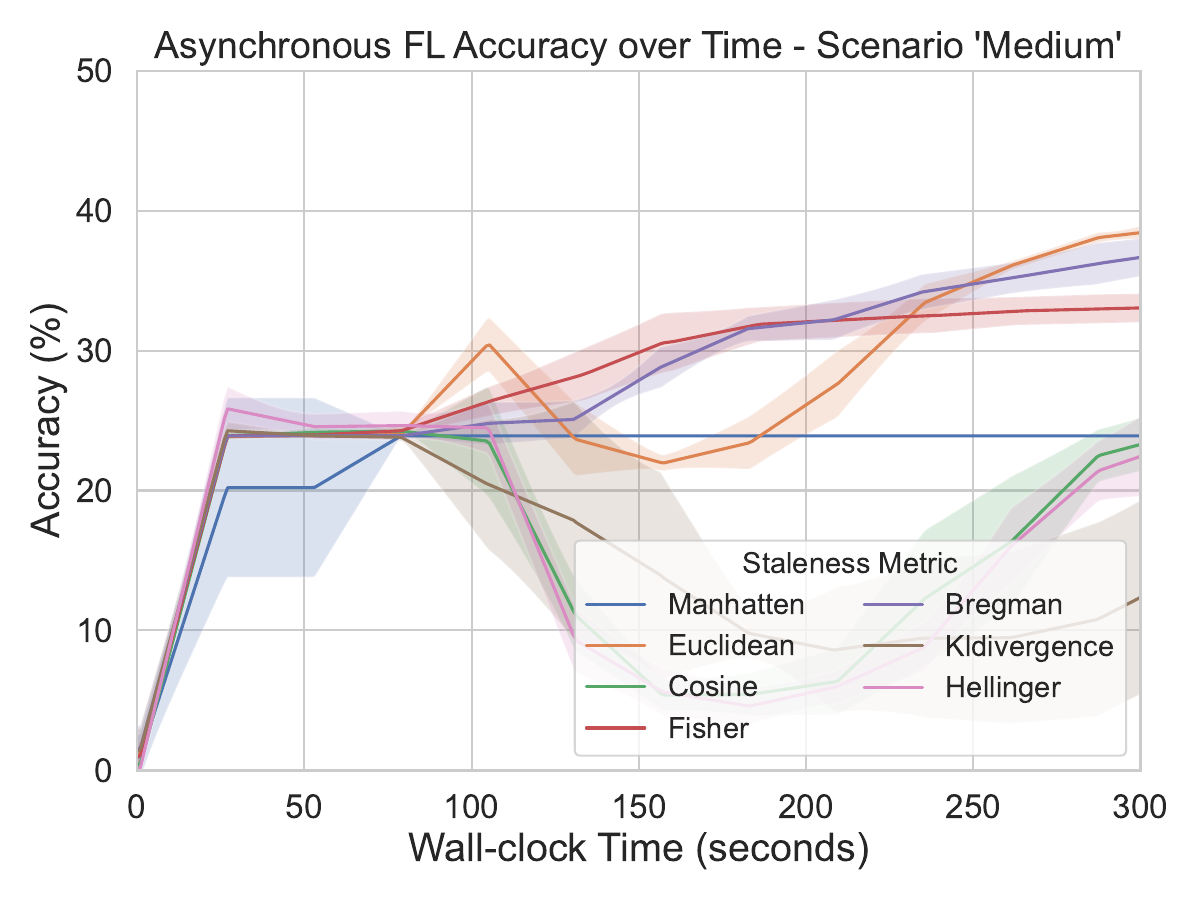}
\caption{Next Character Prediction using an LSTM Model. Test accuracy over wall-clock time for the \textit{Medium} client heterogeneity scenario.}
\label{fig:lstm_accuracy_medium}
\end{figure}

\begin{figure}[h]
\centering
\includegraphics[width=0.5\textwidth]{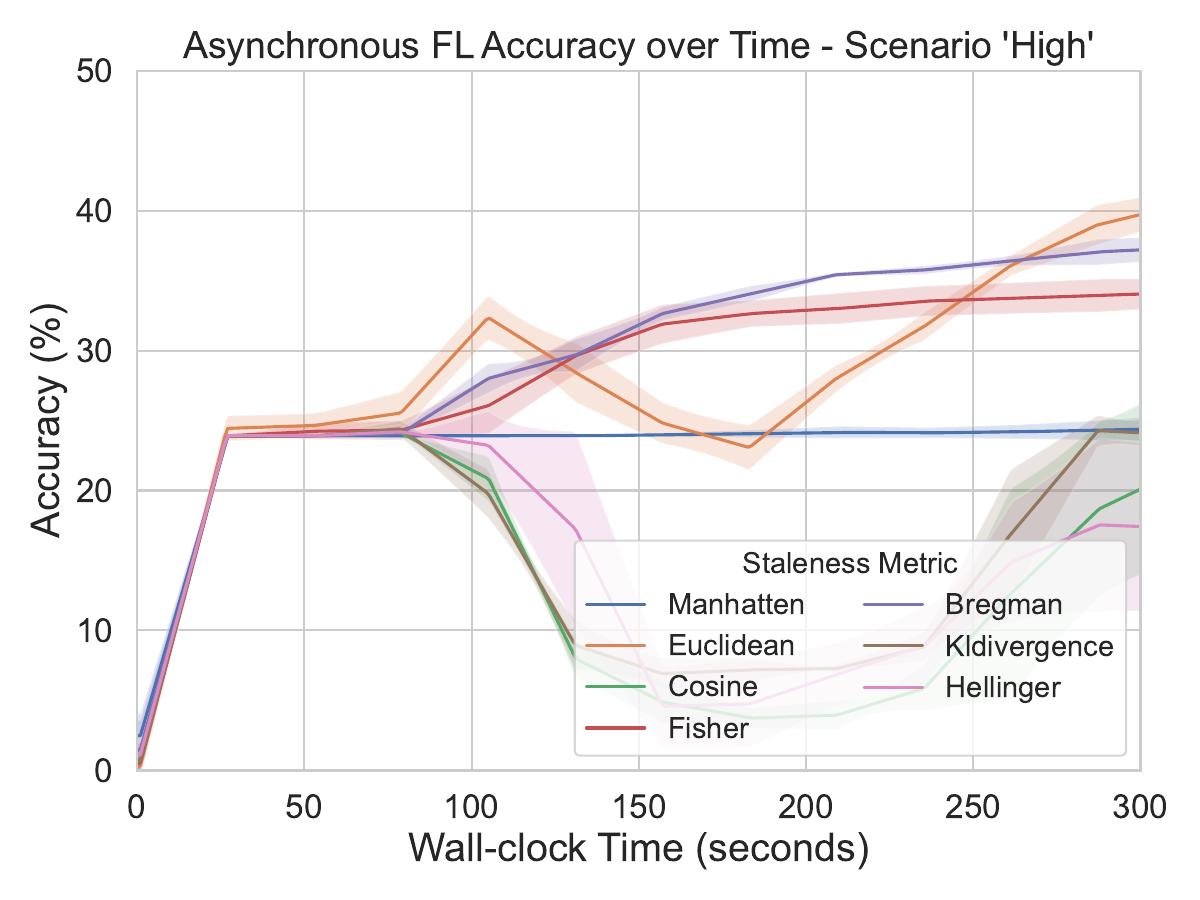}
\caption{Next Character Prediction using an LSTM Model. Test accuracy over wall-clock time for the \textit{High} client heterogeneity scenario.}
\label{fig:lstm_accuracy_high}
\end{figure}

For the next-character prediction task using an LSTM model, the behavior of the various distance metrics diverged notably over all scenarios as it can be seen in Figure \ref{fig:lstm_accuracy_low}, \ref{fig:lstm_accuracy_medium}, \ref{fig:lstm_accuracy_high}. Bregman divergence consistently demonstrated the most stable convergence and achieved the highest overall accuracy across all scenarios, closely followed by the Euclidean distance. While the Euclidean metric reached comparable performance levels, its convergence was notably less stable, exhibiting intermittent drops in model accuracy throughout training. Similar instability was observed with the Cosine, Hellinger, and KL divergence metrics, particularly between 100 and 300 seconds, where their accuracy declined sharply. This suggests that these metrics struggle to robustly accommodate asynchronous updates in a federated learning context, only stabilizing around the 250-second mark.

Notably, the Manhattan distance metric exhibited surprising simplicity and robustness. It was the only metric to converge within the first 50 seconds and remained stable across all scenarios. In contrast, while Bregman and Fisher distances showed continuous improvement over extended computation time, their early-stage convergence was slower.

In summary, even when applied to different tasks and model architectures, Bregman divergence consistently provided the most reliable and often highest-performing measure of gradient staleness, supporting more effective handling of asynchronous updates in federated learning.

\section{Discussion}
\label{sec:discussion}

\textbf{Staleness-Aware Metrics in Practice.}
Our results demonstrate that not all distance metrics are equally effective in mitigating gradient staleness. While Euclidean distance remains a popular choice due to its simplicity and geometric interpretability, Bregman divergence consistently outperformed all other metrics across all availability scenarios.

\textbf{Why Does Bregman Work Better?}
The Bregman divergence generalizes distance through a convex generator function $\phi(x)$. When $\phi(x) = \frac{1}{2} \|x\|^2$, Bregman reduces to the squared Euclidean distance:
\[
D_{\phi}(x, y) = \phi(x) - \phi(y) - \langle \nabla \phi(y), x - y \rangle.
\]
Unlike the symmetric Euclidean distance, Bregman divergence captures directional deviation, which is particularly relevant in asynchronous FL where updates may be outdated by varying degrees. This asymmetry allows it to penalize stale gradients more accurately by modeling their informational deviation from the current model trajectory.

\textbf{Metric Families and Performance.}
While geometric distances such as Manhattan and Euclidean generally performed well, information-theoretic metrics like Kullback-Leibler and Hellinger divergence suffered from high variance and instability, likely due to their sensitivity to small parameter shifts and asymmetric distributions under non-IID data. Fisher information distance emerged as a competitive alternative, suggesting it may be particularly suited for high-staleness regimes.

\textbf{Computational Considerations.}
Although not the focus of this study, certain metrics (e.g., KL-divergence, Fisher) entail higher computational costs. Future work should quantify the trade-off between metric complexity and training gains, especially in resource-constrained edge environments.

\textbf{Toward Practical Deployment.}
From a systems perspective, our findings suggest that asynchronous FL frameworks should expose staleness-handling as a modular component, allowing practitioners to select or tune divergence metrics based on deployment scenarios. Integrating efficient Bregman-based aggregation strategies could improve convergence without additional communication or system overhead.

These observations motivate several promising directions, including dynamic metric selection, layer-wise staleness handling, and adaptive weighting schemes. All of which could make AFL more robust and adaptable in real-world, heterogeneous deployments.

\section{Conclusion}
\label{sec:conclusion}

In this work, we investigated the role of distance metrics in quantifying gradient staleness within asynchronous Federated Learning (AFL). While existing methods such as AsyncFedED rely on simple geometric distances like the Euclidean norm, our study demonstrates that the choice of staleness metric significantly affects convergence behavior, training stability, and final model performance.

Through experiments across varying levels of client heterogeneity and diverse tasks,including computer vision and text prediction,we find that different metrics exhibit distinct behaviors depending on the application domain. For example, Bregman divergence consistently outperforms other metrics in vision tasks, yielding higher accuracy and faster convergence, whereas certain tasks like text prediction benefit more from alternative measures. Moreover, more complex divergence-based metrics such as Kullback-Leibler and Hellinger tend to suffer from high variance and poor early performance, particularly under high staleness conditions.

Our results underscore that staleness is a multifaceted phenomenon that cannot be adequately captured by a single scalar metric across all scenarios. Selecting appropriate distance measures tailored not only to system and data heterogeneity but also to the specific task can significantly enhance the robustness and effectiveness of asynchronous FL systems.

These insights lay the groundwork for more adaptive and context-aware staleness handling mechanisms in future AFL frameworks. In particular, they open up promising avenues for meta-systems that automatically select or adapt staleness metrics based on the application domain, bringing us closer to deploying federated learning reliably in real-world, heterogeneous environments.

\vspace{12pt}

\end{document}